\documentclass[10pt,letterpaper]{article}
\usepackage[top=0.85in,left=2.75in,footskip=0.75in,marginparwidth=2in]{geometry}
\usepackage[utf8]{inputenc}
\usepackage{cite}
\usepackage{nameref,hyperref}

\usepackage{microtype}
\DisableLigatures[f]{encoding = *, family = * }
\usepackage{changepage}
\usepackage[aboveskip=1pt,labelfont=bf,labelsep=period,singlelinecheck=off]{caption}
\usepackage{lastpage,fancyhdr,graphicx,epstopdf}
\usepackage{xcolor,sidecap,wrapfig}
\usepackage[pscoord]{eso-pic}
\usepackage[fulladjust]{marginnote}
\usepackage{amsmath,amssymb,amsthm,booktabs}
\usepackage{tikz}
\usetikzlibrary{arrows.meta,positioning}

\raggedright
\definecolor{Gray}{gray}{.25}
\reversemarginpar
\fancyheadoffset[L]{2.25in}
\fancyfootoffset[L]{2.25in}
\makeatletter
\renewcommand{\@biblabel}[1]{\quad#1.}
\makeatother

\begin{document}
\vspace*{0.35in}
\begin{flushleft}
{\Large
\textbf{CXR-CMFA: A Cross-Modal Feature Aggregation Network for Chest X-Ray Pathology Classification}
}
\newline
\\
Talukder Hasnat Zadid\textsuperscript{1,*},
Jahin Sultana\textsuperscript{1},
Felipe Bochini Brisola\textsuperscript{2},
Kishore Roy\textsuperscript{1},
Mysha Rahman\textsuperscript{1},
Syeda Asrafa Islam\textsuperscript{1},
Md. Abdullah-Al-Jubair\textsuperscript{1}
\\
\bigskip
\bf{1} American International University--Bangladesh (AIUB), Dhaka, Bangladesh
\\
\bf{2} University of Sorocaba (UNISO), São Paulo, Brazil
\\
\bigskip
* 22-46808-1@student.aiub.edu
\end{flushleft}

\section*{Abstract}
Chest X-ray (CXR) interpretation remains challenging because subtle pathology can be obscured by overlapping anatomy, image-quality variation, and limited visual evidence in isolated radiographs.
Although image-only deep learning models provide useful baselines, they discard the complementary diagnostic context available in radiology reports, while large vision-language models can be computationally expensive for practical deployment.
This paper presents \textbf{CXR-CMFA}, a lightweight cross-modal feature aggregation network for binary CXR pathology classification.
The proposed framework combines a dual-stream visual encoder, using EfficientNet-B0 for global anatomical representation and DenseNet-121 for local pathological detail, with a TF-IDF--MLP report encoder.
Two fusion modules integrate these modalities: the Dual-Modal Spatial Attention Module (DMSAM) uses the report embedding to gate global and local visual features, and the Cross-Modal Gated Feature Fusion (CGFF) module learns per-instance weights for visual and textual contributions.
Experiments on the MIMIC-CXR benchmark, using the provided train, validation, and test splits, show that CXR-CMFA achieves 93.45\% accuracy, 99.88\% precision, 91.8\% recall, 95.7\% F1-score, and 99.19\% AUC on the held-out test set.
The model produces only four false positives among 984 normal test cases, supporting its suitability for high-specificity screening workflows.
An ablation study further demonstrates a 22.47 percentage-point AUC improvement over the global-image-only variant, confirming the value of adaptive cross-modal fusion.
These results indicate that compact report-conditioned fusion can provide strong discriminative performance without relying on large pretrained language models.
\medskip

\noindent\textbf{Keywords:} Cross-Modal Fusion, Chest X-Ray Classification, Multimodal Learning, Medical Image Analysis, Radiology Report Encoding, EfficientNet-B0, DenseNet-121, TF-IDF, Attention Mechanism, MIMIC-CXR, Pathology Detection, Lightweight Vision-Language Model

\section*{Introduction}

Advancements in Artificial Intelligence (AI) have driven substantial progress in automated disease identification from chest X-rays (CXR), supporting pathology classification and intelligent clinical decision support~\cite{ref5,ref8,ref7}. Despite all these innovations, CXR continues to be a challenging analysis and interpretation task due to overlapping anatomical structures, variable image quality, subtle radiographic findings, and the large number of studies to be analysed by experts~\cite{ref5}. Thus, automated, reliable analysis of CXR is not only a technical challenge, but an important clinical and public-health goal.

These requirements are now fulfilled by employing multimodal approaches, where both visual and textual data are used to enhance medical image analysis, as reported in~\cite{ref3,biovil2023}. There are, however, many systems already in use that are image-only or use relatively simple fusion techniques, which do not take full advantage of the complementary diagnostic context encapsulated in the free-text radiology report. Thus, there is a need for efficient, clinically useful multimodal schemes that can embed CXR images with clinical textual data.

CheXNet provides a strong image-only CNN baseline~\cite{ref8}. Nasser and Akhloufi~\cite{ref5} confirmed the continuing importance of CNNs for chest-disease detection, while Hasanah et al.~\cite{ref8} added an FPN for multi-scale, multi-label classification. Nevertheless, image-only models cannot access the diagnostic context in radiology reports.

Transformer-based vision models have recently emerged as alternatives to the conventional CNNs. DAViT is an approach by Fu et al.~\cite{ref7} that leverages the self-attention mechanism to detect pneumonia in chest X-rays, showcasing the promise of self-attention for capturing hierarchical relationships in medical images. A similar investigation of the efficient representation of medical images for classification using transformers is detailed in MedViT~\cite{medvit2023}. However, transformer models can be resource-intensive and require extensive pretraining, potentially hindering their adoption in low-resource healthcare settings.

Multimodal approaches provide a means of combining complementary visual and textual information. Both TieNet~\cite{tienet2018} and Drozdov et al.~\cite{ref10} showed that it is possible to jointly use the image and report information to create a text-image embedding framework, and that language models can be trained to learn useful clinical representations directly from the radiology-report text. Recent biomedical vision-language methods also illustrate the benefits of representation alignment between images and texts for medical-image understanding~\cite{biovil2023}. Ullah et al.~\cite{ref3} proposed a multimodal model for medical image analysis using class-specific feature selection, composed of two streams. Recently, Lee et al.~\cite{ref6} introduced a multimodal large language model (LLM), named CXR-LLaVA, for chest X-ray image interpretation. Despite its impressive architecture, which is based on significantly larger pretrained models than the lightweight approach presented here, the capacity of CXR-LLaVA for large-scale vision-language alignment is highlighted.

{Nevertheless, many multimodal systems use fixed or globally learned fusion strategies that do not explicitly account for the reliability of each modality for an individual input. Furthermore, large vision-language models can introduce considerable computational overhead. Prior modality-aware fusion approaches such as MAINet}~\cite{wei2025mainet} {demonstrate the importance of explicitly modeling interactions between modalities, but the combination of text-conditioned dual-stream visual attention and per-instance adaptive modality gating remains insufficiently explored for CXR classification.}

This paper proposes \textbf{CXR-CMFA} (Chest X-Ray Cross-Modal Fusion Architecture), a multimodal framework that addresses these limitations through three tightly integrated components.

First, a \textit{dual-stream visual encoder} captures both coarse anatomical context (global 224$\times$224 view via EfficientNet-B0 with Squeeze-and-Excitation recalibration) and fine-grained pathological detail (local 96$\times$96 patch via DenseNet-121).  {EfficientNet-B0 provides an efficient convolutional backbone} ~\cite{tan2019efficientnet}, {while Squeeze-and-Excitation blocks improve channel-wise feature recalibration}~\cite{hu2018senet}.

Second, the \textit{Dual-Modal Spatial Attention Module} (DMSAM) uses the TF-IDF report embedding to independently gate each visual stream before merging them. {TF-IDF provides a lightweight representation that emphasizes diagnostically informative terms}~\cite{wei2025mainet}, {while attention-based representations provide an established mechanism for emphasizing informative components of sequential text}~\cite{yang2016hierarchical}. {This allows clinically salient report information to influence both global and local visual representations.}

Third, the \textit{Cross-Modal Gated Feature Fusion} (CGFF) mechanism learns per-instance scalar gates $(\alpha, \beta, \gamma)$ to adaptively weight the visual fusion output, the local visual stream, and the textual embedding. {The mechanism therefore enables the model to down-weight the report when it provides limited information and increase its contribution when it contains strong diagnostic signals.}

Unlike large biomedical vision-language models such as BioViL~\cite{biovil2023} or CXR-LLaVA~\cite{ref6}, CXR-CMFA encodes reports through a lightweight TF-IDF pipeline. { This design reduces the dependence on large pretrained language models while retaining clinically meaningful textual information.}

Our ablation demonstrates a \textbf{22.47 percentage-point AUC improvement} over the global-image-only variant (76.58\% $\to$ 99.05\%), quantifying the contribution of cross-modal fusion. In the primary held-out test evaluation, CXR-CMFA achieves 99.19\% AUC and produces only \textbf{4 false positives} out of 984 normal cases, corresponding to a specificity of 99.59\% and supporting its potential for high-precision screening.

\noindent The main contributions of this paper are as follows:
\begin{itemize}

\item \textbf{Dual-Stream Visual Encoding:} A two-branch visual encoder jointly captures global anatomical structure and local pathological texture from CXR images, with channel attention via SE blocks.

\item \textbf{Dual-Modal Spatial Attention Module (DMSAM):} A text-conditioned visual attention mechanism that independently gates the global and local visual streams using the report embedding before merging them, enabling semantically guided feature discrimination.

\item \textbf{Cross-Modal Gated Feature Fusion (CGFF):} An adaptive per-instance gating network that learns modality-specific contribution weights $(\alpha, \beta, \gamma)$, allowing the model to flexibly balance visual and textual streams according to the informativeness of each input.

\item \textbf{Empirical validation:} Experiments on MIMIC-CXR demonstrate 99.88\% precision and 0.9919 AUC, while the ablation protocol shows a 22.47 percentage-point AUC gain over the global-image-only variant, outperforming image-only baselines across all five metrics and achieving the highest precision among the compared models.
\end{itemize}

The remainder of this paper is organised as follows. Section~\nameref{sec:related} reviews related work on CXR classification and multimodal learning. Section~\nameref{sec:method} details the proposed CXR-CMFA framework. Section~\nameref{sec:results} presents experimental results and ablation analysis. Section~\nameref{sec:discussion} discusses findings and limitations. Section~\nameref{sec:conclusion} concludes and outlines future directions.

\section*{Related Work}
\label{sec:related}

\subsection*{Image-Only CXR Classification}

CheXNet uses a DenseNet-121 backbone for image-only pneumonia detection~\cite{ref8}. Nasser and Akhloufi~\cite{ref5} emphasized the importance of CNN representations, while Hasanah et al.~\cite{ref8} enhanced CheXNet with an FPN for multi-label classification. However, image-only models omit report context and achieve AUC values around 74--76\% on MIMIC-CXR.

Transformer-based visual architectures have subsequently been investigated for medical-image analysis.{ DAViT uses domain adaptation and transformer-based attention to model long-range relationships in chest radiographs}~\cite{ref7}, {while MedViT explores robust and efficient transformer representations for medical-image classification}~\cite{medvit2023}. However, transformer-based systems may require substantial computational resources and pretraining data, motivating lightweight alternatives for resource-constrained environments.

\subsection*{Vision-Language Models for Radiology}

CXR images have been used frequently in studies to address this problem, but more recently, researchers have been incorporating the radiology report with the CXR image. In order to jointly model visual and textual representations, {TieNet proposed a text-image embedding architecture, in which the image is represented by a text embedding and a text image embedding is jointly generated by attention mechanisms}~\cite{tienet2018}. Drozdov et al.~\cite{ref10} further demonstrated that language models can learn useful clinical representations directly from radiology-report text, highlighting the diagnostic information present in the textual modality.

{Biomedical vision-language research has further established the value of aligning medical images and text representations. BioViL-T, for example, jointly models biomedical images and text while incorporating temporal information from clinical imaging sequences}~\cite{biovil2023}. These approaches demonstrate that textual information can provide complementary clinical context beyond purely visual representations.

{More recently, Lee et al.}~\cite{ref6} {introduced CXR-LLaVA, a multimodal large language model for CXR interpretation. The model combines a vision encoder with a large language model and demonstrates the potential of large-scale multimodal architectures for chest radiography.}However, such systems rely on considerably larger pretrained components than the lightweight text representation used in CXR-CMFA.

Ullah et al.~\cite{ref3} also demonstrated the effectiveness of dual-stream multimodal medical-image analysis with class-specific feature selection. These studies collectively motivate the use of visual and textual modalities while leaving open the question of how modality contributions should be adapted for individual CXR cases.

\subsection*{Multi-Modal Fusion Strategies}

The way the visual and text-based representations are combined has significant effects on multimodal performance. {In early multimodal approaches, image and text features were simply concatenated, without explicitly modelling higher-order cross-modal interactions.} {A more expressive mechanism to model interaction between modalities was introduced in Multimodal Compact Bilinear Pooling, but this leads to a higher computational burden}~\cite{fukui2016mcb}.

Attention-based architectures provide another strategy for learning relationships between modalities. CrossViT demonstrated the use of {cross-attention mechanisms for exchanging information between different visual representations}~\cite{chen2021crossvit}. Similarly, attention-based language models have demonstrated the ability to emphasize informative portions of textual inputs~\cite{yang2016hierarchical}. However, conventional attention mechanisms do not necessarily account for the fact that the diagnostic reliability of a radiology report can vary substantially from one case to another.

{A CXR report may be short, incomplete, or contain limited discriminative information. Consequently, assigning a fixed contribution to the textual modality may be suboptimal.} {MAINet}~\cite{wei2025mainet} {demonstrates the importance of explicitly modeling modality-aware interactions, while CXR-CMFA extends this idea by introducing explicit per-instance gating.}

CXR-CMFA addresses these limitations through two complementary components. First, the Dual-Modal Spatial Attention Module (DMSAM) uses the report embedding as an attention prior over visual features \emph{independently} for the global and local views. {This allows textual information to guide visual feature extraction without requiring a symmetric image-text alignment. Second, the Cross-Modal Gated Feature Fusion (CGFF) mechanism learns per-instance scalar gates} $(\alpha, \beta, \gamma)$ {to adaptively weight the fused visual representation, local visual representation, and textual representation.}

This design enables the model to reduce the contribution of an uninformative report while increasing the contribution of diagnostically useful textual information. The combination of dual-stream text-conditioned visual attention and instance-wise modality gating therefore provides a lightweight alternative to large vision-language architectures.

\subsection*{Explainability and Clinical Interpretability}

{Explainability is an important consideration for medical AI because high predictive performance alone does not guarantee that model decisions are clinically interpretable. Recent work has examined explainable AI techniques such as SHAP and LIME for understanding feature contributions and improving the transparency of machine-learning systems}~\cite{ahmed2026explainable}. In the context of CXR analysis, spatial attention mechanisms can additionally provide information about which image regions contribute to the model's prediction.

The proposed CXR-CMFA architecture is designed with interpretability in mind by explicitly separating global and local visual representations and by conditioning visual attention on textual information. This structure provides a more transparent representation of how image and report information interact than a single opaque multimodal embedding.

The literature reveals three persistent gaps: image-only models discard report context, conventional fusion often lacks per-instance adaptation, and large vision-language models are computationally expensive. CXR-CMFA addresses these limitations through dual-stream visual encoding, text-conditioned attention, instance-wise fusion gates, and a lightweight TF-IDF report representation. The following sections describe the approach and its evaluation on MIMIC-CXR.

\section*{Methodology}
\label{sec:method}

Figure~\ref{fig:methodology_flow} presents the proposed CXR-CMFA methodology, beginning with CXR/report inputs and proceeding through preprocessing, dual-stream feature extraction, report-conditioned DMSAM fusion, CGFF adaptive aggregation, and final binary prediction.

\subsection*{Dataset}

\begin{figure}[h]
  \centering
  \includegraphics[width=\columnwidth]{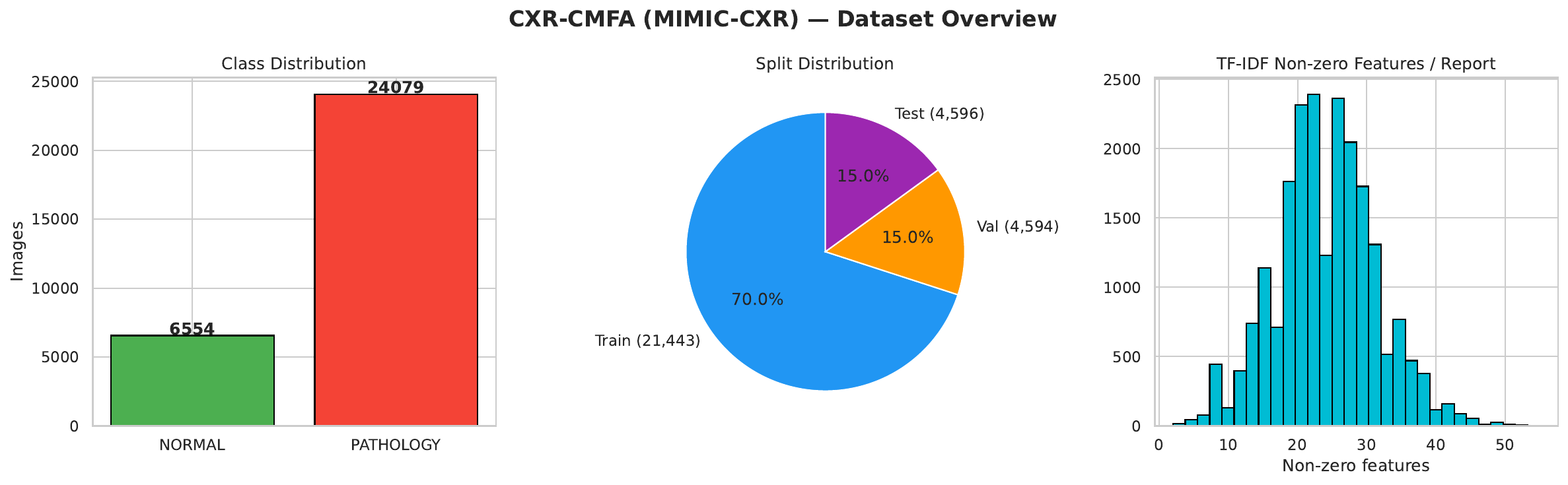}
  \caption{MLforHealthcare/MIMIC-CXR dataset overview: predefined train/validation/test split sizes and histogram of non-zero TF-IDF features per report.}
  \label{fig:dataset}
\end{figure}

\begin{figure}[h]
  \centering
  \includegraphics[width=\columnwidth]{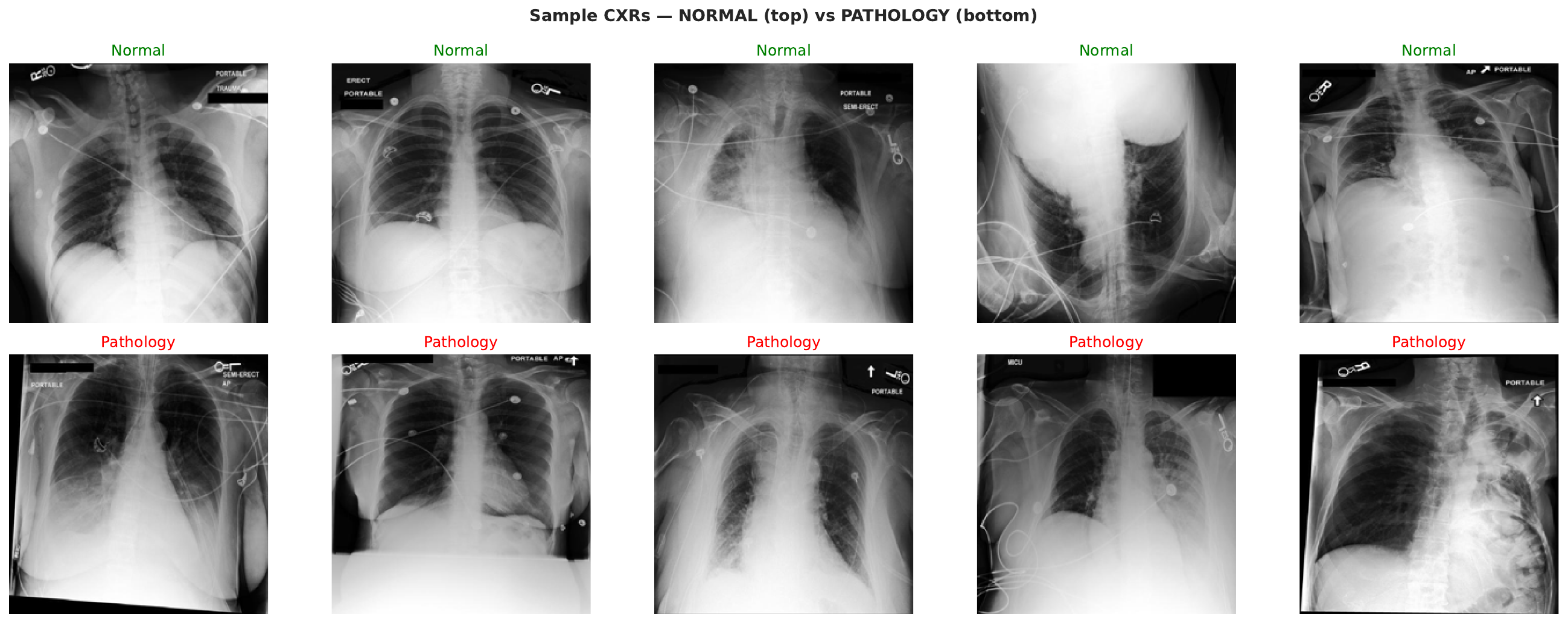}
  \caption{Representative CXR images from the dataset. Top row: NORMAL cases; bottom row: PATHOLOGY cases. Visual similarity between classes motivates complementary report-text features.}
  \label{fig:samples}
\end{figure}

We use the \textit{MLforHealthcare/MIMIC-CXR} dataset\footnote{\textit{MLforHealthcare/mimic-cxr}. Hugging Face. \url{https://huggingface.co/datasets/MLforHealthcare/mimic-cxr}. Accessed 6 June 2026.}. This machine-learning-ready subset is derived from the MIMIC-CXR database~\cite{johnson2019mimic} and comprises 30,633 chest X-ray image--report pairs.
The dataset is organized into three predefined splits: \textit{train} (21,443 image--report pairs), \textit{validation} (4,594 image--report pairs), and \textit{test} (4,596 image--report pairs).
Each sample consists of a CXR image (\texttt{image}) and a radiology report (\texttt{reports}); the image entries are provided at 512$\times$512 pixel resolution.
Figure~\ref{fig:dataset} summarizes the provided split sizes and the distribution of non-zero TF-IDF features per report.
Figure~\ref{fig:samples} illustrates representative CXR images from the dataset.

\subsection*{Label Assignment}

Binary labels are derived programmatically from the supplied report field.
A study is assigned NORMAL (0) if the impression contains negation phrases such as \textit{``no acute finding''} or \textit{``no acute cardiopulmonary abnormality''}; otherwise, a curated keyword list spanning 17 pathological terms (pneumonia, pneumothorax, effusion, consolidation, opacity, atelectasis, cardiomegaly, edema, infiltrate, nodule, mass, fracture, lesion, enlarged, congestion, pleural, haziness) is matched with word-boundary regular expressions.
Any match yields PATHOLOGY (1).
Reports without any matched pathology keyword are assigned NORMAL; empty or very short strings, if encountered, are handled by the preprocessing fallback below.

\subsection*{Text Preprocessing}

Raw report text undergoes the following pipeline: (i) lowercasing and whitespace normalisation; (ii) removal of non-alphanumeric characters except punctuation used in clinical abbreviations; (iii) stopword removal.
A minimum report length of five characters is enforced; shorter strings are replaced with the token sequence \texttt{no acute cardiopulmonary abnormality}.

\subsection*{Feature Engineering}

Report text is encoded using \textbf{TF-IDF} (Term Frequency–Inverse Document Frequency) with the following configuration: vocabulary capped at 128 features, unigram and bigram ($n$-gram range $[1,2]$), minimum document frequency of 2, sublinear TF scaling ($\log(1+\text{tf})$), and $\ell_2$ normalisation of the resulting vectors.
This produces a compact, interpretable 128-dimensional report embedding that is efficient to compute and does not require a large pretrained language model at inference time.
The distribution of non-zero features per report shown in Figure~\ref{fig:dataset} confirms that most reports activate 15--35 features, providing a sufficiently dense signal for the report encoder.

\subsection*{Data Splitting}

{We use the dataset's predefined splits rather than creating a new random partition: 21,443 training image--report pairs, 4,594 validation image--report pairs, and 4,596 test image--report pairs. All TF-IDF vocabulary and normalisation statistics are fitted exclusively on the training set and applied to validation and test sets to prevent data leakage.}

\subsection*{CXR-CMFA Architecture}

\begin{figure*}[t]
  \centering
  \resizebox{0.94\textwidth}{!}{%
  \begin{tikzpicture}[
    font=\sffamily,
    >=Stealth,
    node distance=1.35cm and 1.45cm,
    input/.style={draw, rounded corners=5pt, line width=0.8pt, fill=blue!8, align=center, minimum width=2.7cm, minimum height=1.25cm},
    encoder/.style={draw, rounded corners=5pt, line width=0.8pt, fill=green!9, align=center, minimum width=2.55cm, minimum height=1.15cm},
    fusion/.style={draw, rounded corners=6pt, line width=0.9pt, fill=orange!14, align=center, minimum width=2.9cm, minimum height=1.25cm},
    output/.style={draw, rounded corners=5pt, line width=0.8pt, fill=red!8, align=center, minimum width=2.65cm, minimum height=1.15cm},
    flow/.style={->, line width=0.85pt},
    skip/.style={->, line width=0.75pt, dashed, draw=gray!75}
  ]
    \node[input] (cxr) at (0,2.25) {\Large$\square$\\[-1pt]\textbf{Full CXR}\\\footnotesize 224$\times$224};
    \node[input] (crop) at (0,0) {\Large$\boxplus$\\[-1pt]\textbf{Local Crop}\\\footnotesize 96$\times$96};
    \node[input] (report) at (0,-2.25) {\Large$\mathcal{R}$\\[-1pt]\textbf{Report Text}};

    \node[encoder, right=of cxr] (global) {\textbf{Global}\\Encoder};
    \node[encoder, right=of crop] (local) {\textbf{Local}\\Encoder};
    \node[encoder, right=of report] (text) {\textbf{TF--IDF}\\+ MLP};

    \node[fusion, right=1.55cm of local] (dmsam) {\textbf{DMSAM}\\\footnotesize report-gated\\\footnotesize visual fusion};
    \node[fusion, right=1.7cm of dmsam] (cgff) {\textbf{CGFF}\\\footnotesize adaptive gates\\$\alpha,\beta,\gamma$};
    \node[output, right=1.55cm of cgff] (classifier) {\textbf{Classifier}\\MLP};
    \node[output, right=1.25cm of classifier] (prediction) {\textbf{Prediction}\\\footnotesize NORMAL/PATHOLOGY};

    \draw[flow] (cxr) -- (global);
    \draw[flow] (crop) -- (local);
    \draw[flow] (report) -- (text);

    \draw[flow] (global.east) -- (dmsam.north west);
    \draw[flow] (local) -- (dmsam);
    \draw[flow] (text.east) -- (dmsam.south west);

    \draw[flow] (dmsam) -- (cgff);
    \draw[skip] (local.north east) to[out=25,in=155] (cgff.north west);
    \draw[skip] (text.north east) to[out=30,in=-155] (cgff.south west);

    \draw[flow] (cgff) -- (classifier);
    \draw[flow] (classifier) -- (prediction);
  \end{tikzpicture}%
  }
  \caption{Overview of the proposed CXR-CMFA framework. The full CXR image, local image crop, and radiology report are processed by separate branches. Global and local visual features are extracted from the images, while the report is converted into a text embedding. DMSAM fuses the visual features using report guidance, and CGFF combines visual and text information before final classification. Dashed arrows indicate skip connections to CGFF.}
  \label{fig:methodology_flow}
\end{figure*}
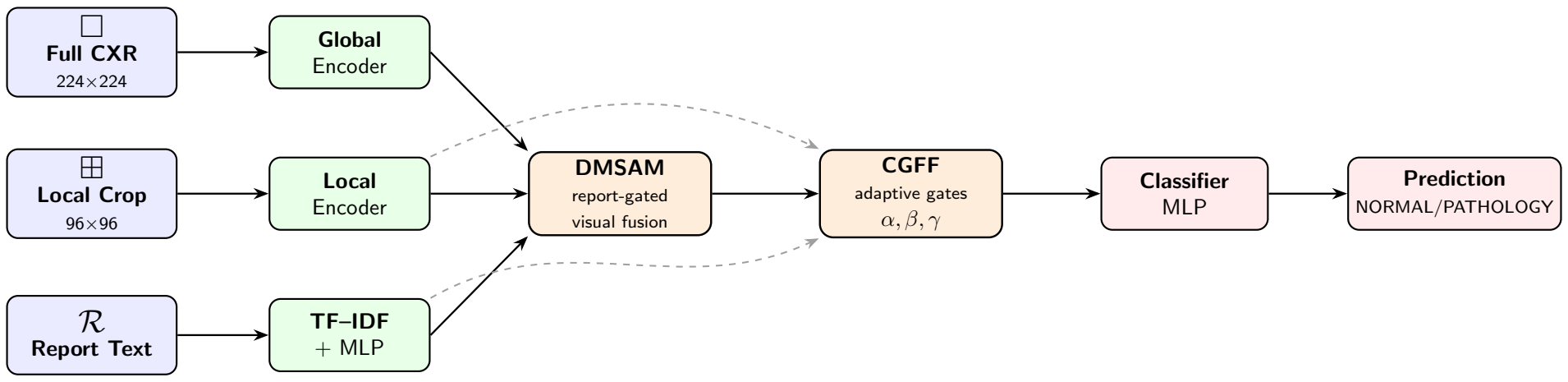

\begin{figure}[t]
  \centering
  \includegraphics[width=\columnwidth]{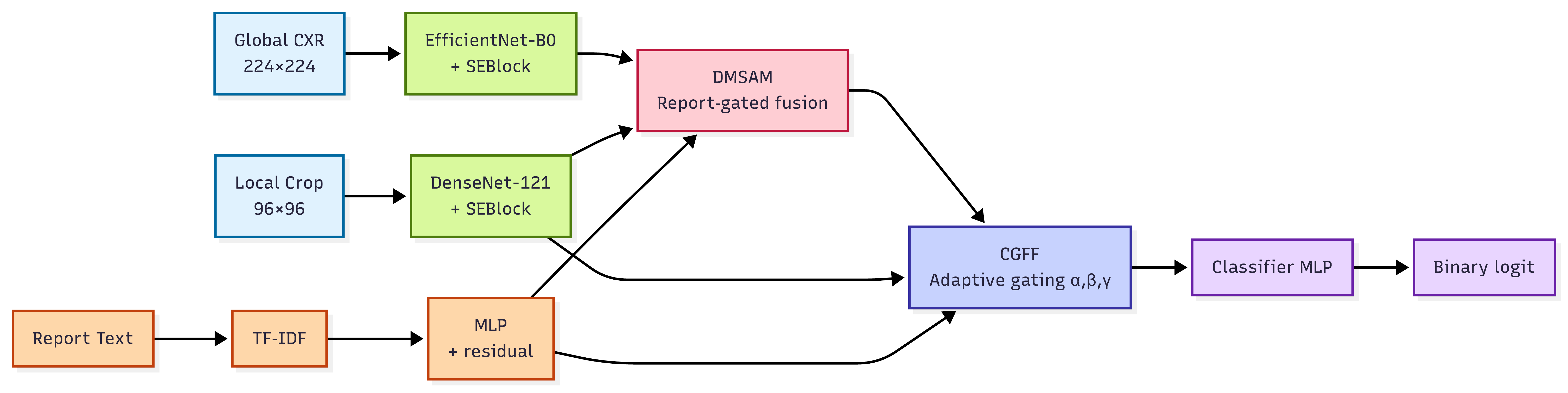}
  \caption{Detailed CXR-CMFA model architecture. The full CXR image, local crop, and radiology report are encoded through separate visual and textual branches. The DMSAM block performs report-conditioned fusion of global and local visual features, and the CGFF block adaptively aggregates the fused visual representation with local visual and textual features before binary classification.}
  \label{fig:model_architecture}
\end{figure}

The model components in Figure~\ref{fig:methodology_flow} show how CXR-CMFA implements the proposed methodology, while Figure~\ref{fig:model_architecture} presents the detailed model architecture.
The model receives three complementary inputs: a full 224$\times$224 CXR image for global anatomical context, a 96$\times$96 local crop for fine-grained pathology cues, and the corresponding radiology report represented with TF-IDF features.
The global image stream is processed by EfficientNet-B0 with an SE block, the local stream is processed by DenseNet-121 with an SE block, and the report stream is encoded by an MLP with a residual connection.
These representations are combined in two stages: DMSAM first performs report-gated visual fusion, and CGFF then learns adaptive modality weights $(\alpha, \beta, \gamma)$ before a classifier MLP produces the final binary logit.
The flowchart emphasises the complete methodological path from multimodal inputs to the final NORMAL/PATHOLOGY prediction, whereas the architecture diagram highlights the concrete network modules and feature-aggregation route.

\paragraph{Global Encoder ($\mathcal{E}_g$).}
A pre-trained EfficientNet-B0~\cite{tan2019efficientnet} processes the full-resolution 224$\times$224 CXR.
The original classifier head is replaced with: a Squeeze-and-Excitation (SE) block for channel-wise recalibration~\cite{hu2018senet} followed by a projection layer (Linear--BN--ReLU--Dropout$_{0.4}$) mapping to $\mathbb{R}^{512}$.
EfficientNet-B0's compound scaling provides an efficient balance between receptive field coverage and parameter count.

\paragraph{Local Encoder ($\mathcal{E}_l$).}
A pre-trained DenseNet-121~\cite{ref8} processes a 96$\times$96 random crop (training) or centre crop (inference) of the same image at the 224$\times$224 intermediate resolution.
This local view captures fine-grained textures (e.g., subtle interstitial opacities, early nodules) that may be suppressed in the global downsampled view.
An identical SE + projection block maps the output to $\mathbb{R}^{512}$.

\paragraph{Report Encoder ($\mathcal{E}_r$).}
The 128-d TF-IDF vector passes through a two-layer MLP (128$\to$256$\to$256-d) with BatchNorm, ReLU, and Dropout, augmented by a residual skip connection from input to output.
This produces a 256-d linguistic embedding $\mathbf{f}_r \in \mathbb{R}^{256}$.

\paragraph{Dual-Modal Spatial Attention Module (DMSAM).}
Given global embedding $\mathbf{f}_g$, local embedding $\mathbf{f}_l$, and report embedding $\mathbf{f}_r$, DMSAM first projects $\mathbf{f}_r$ into the respective visual spaces:
\begin{equation}
  \mathbf{a}_g = \sigma\!\left(\mathbf{W}_{rg}\,\mathbf{f}_r\right), \quad
  \mathbf{a}_l = \sigma\!\left(\mathbf{W}_{rl}\,\mathbf{f}_r\right)
\end{equation}
These are used as soft attention masks on each visual stream:
\begin{equation}
  \tilde{\mathbf{f}}_g = \mathbf{f}_g \odot \sigma\!\left(\mathbf{W}_{gg}(\mathbf{f}_g \odot \mathbf{a}_g)\right), \quad
  \tilde{\mathbf{f}}_l = \mathbf{f}_l \odot \sigma\!\left(\mathbf{W}_{ll}(\mathbf{f}_l \odot \mathbf{a}_l)\right)
\end{equation}
A learnable scalar $w \in (0,1)$ balances the two streams, and the attended representations are projected via:
\begin{equation}
  \mathbf{f}_v = \text{BN-ReLU}\!\left(\mathbf{W}_f\,[w\,\tilde{\mathbf{f}}_g \;\|\; (1{-}w)\,\tilde{\mathbf{f}}_l]\right) \in \mathbb{R}^{512}
\end{equation}
By conditioning each visual stream's gating function on the report separately before fusion, DMSAM allows the language to steer attention differently in the global and local views.

\paragraph{Cross-modal Gated Feature Fusion (CGFF).}
Given $\mathbf{f}_v$ (DMSAM output), $\mathbf{f}_l$, and $\mathbf{f}_r$, CGFF learns per-instance gates:
\begin{equation}
  [\alpha, \beta, \gamma] = \text{Softmax}\!\left(\mathbf{W}_{\text{gate}}\,[\mathbf{f}_v \;\|\; \mathbf{f}_l \;\|\; \mathbf{f}_r] + \mathbf{b}_{\text{gate}}\right)
\end{equation}
where $\mathbf{W}_{\text{gate}} \in \mathbb{R}^{3 \times 1280}$.
The fused representation is:
\begin{equation}
  \mathbf{f}_{\text{fuse}} = \text{BN-ReLU}\!\left(\alpha\,\mathbf{P}_v\mathbf{f}_v + \beta\,\mathbf{P}_l\mathbf{f}_l + \gamma\,\mathbf{P}_r\mathbf{f}_r\right)
\end{equation}
with projection matrices $\mathbf{P}_v, \mathbf{P}_l, \mathbf{P}_r$ mapping each modality to $\mathbb{R}^{512}$.
Unlike fixed-weight fusion, CGFF adapts to the reliability of each input; if the report is absent or uninformative, the gate automatically down-weights $\gamma$.

\paragraph{Classifier Head.}
$\mathbf{f}_{\text{fuse}} \in \mathbb{R}^{512}$ is passed through a three-layer MLP (512$\to$256$\to$64$\to$1) with BatchNorm, ReLU, and Dropout at each stage, producing a scalar logit for binary classification.

\subsection*{Training Protocol}

All models are trained for up to 10 epochs with early stopping (patience = 4, monitored on validation loss).
The loss function is binary cross-entropy with logits using a class-balanced positive weight $w^+ = N_{\text{neg}}/N_{\text{pos}}$ to counteract the 3.7:1 class imbalance.
Optimisation uses Adam ($\text{lr}=10^{-4}$, $\lambda=10^{-4}$ weight decay) with a ReduceLROnPlateau scheduler (factor 0.5, patience 2).
Input images are augmented during training with random horizontal flip, $\pm10^\circ$ rotation, and colour jitter (brightness and contrast $\pm0.2$).
All experiments use batch size 16, and ImageNet mean/std normalisation.
The random seed is fixed to 42 across all libraries for reproducibility.

\subsection*{Hyperparameter Optimization}

To identify the best configuration for CXR-CMFA, we performed a grid search over key hyperparameters using the validation set.
The search ranges were defined as follows: learning rate $\in \{1\times10^{-4}, 5\times10^{-5}, 1\times10^{-5}\}$, batch size $\in \{8, 16, 32\}$, dropout rate after projection layers $\in \{0.2,0.3,0.4,0.5\}$, weight decay $\in \{1\times10^{-4}, 5\times10^{-5}, 0\}$, and number of epochs $\in \{8,10,12\}$ with early stopping patience $\in \{3,4,5\}$.
Model selection was based on the highest validation area under the ROC curve (AUC).
Table~\ref{tab:hyperparams} reports the final selected values.

\begin{table*}[t]
\centering
\caption{Selected hyperparameters for CXR-CMFA after grid search on the validation set.}
\label{tab:hyperparams}
\small
\resizebox{\textwidth}{!}{%
\begin{tabular}{lcc}
\toprule
\textbf{Parameter} & \textbf{Selected Value} & \textbf{Search Range} \\
\midrule
Learning rate             & $1\times10^{-4}$                         & $\{1\times10^{-5}, 5\times10^{-5}, 1\times10^{-4}\}$ \\
Batch size                & 16                                       & $\{8, 16, 32\}$ \\
Dropout rate (projection) & 0.4                                      & $\{0.2,0.3,0.4,0.5\}$ \\
Weight decay              & $1\times10^{-4}$                         & $\{0, 5\times10^{-5}, 1\times10^{-4}\}$ \\
Number of epochs          & 10                                       & $\{8,10,12\}$ \\
Early stopping patience   & 4                                        & $\{3,4,5\}$ \\
Optimizer                 & Adam                                     & -- \\
Learning rate scheduler   & ReduceLROnPlateau                        & -- \\
Image augmentation        & Random flip, rotation, colour jitter     & -- \\
\bottomrule
\end{tabular}%
}
\end{table*}

All other hyperparameters not listed, such as hidden-layer dimensions and projection sizes, were kept as described in Section~\nameref{sec:method}.
The chosen values yield the fastest convergence with the highest validation AUC while avoiding overfitting.

\section*{Experimental Results and Analysis}
\label{sec:results}

\subsection*{Quantitative Comparison}

Table~\ref{tab:results} reports test-set performance for all five models.
CXR-CMFA achieves the \textbf{highest precision} (99.88\%) and a \textbf{near-top AUC} (99.19\%), surpassing the image-only CheXNet and MedViT-S baselines across all five metrics and exceeding TieNet in precision and AUC.
BioViL records marginally higher accuracy (96.24\% vs.\ 93.45\%) and AUC (99.35\% vs.\ 99.19\%), but at the cost of lower precision (99.65\% vs.\ 99.88\%).

\begin{table}[h]
\caption{{Test-set performance on MIMIC-CXR binary classification. Best value per column in }\textbf{{bold}}{; second-best }\underline{{underlined}}{.}}
\label{tab:results}
\small
\begin{tabular}{lccccc}
\toprule
\textbf{Model} & \textbf{Acc} & \textbf{Prec} & \textbf{Rec} & \textbf{F1} & \textbf{AUC} \\
\midrule
CheXNet~\cite{ref8}  & 0.707 & 0.898 & 0.707 & 0.791 & 0.759 \\
MedViT-S~\cite{medvit2023}           & 0.707 & 0.883 & 0.722 & 0.794 & 0.739 \\
TieNet~\cite{tienet2018}             & \underline{0.960} & \underline{0.995} & \underline{0.954} & \underline{0.974} & \underline{0.990} \\
BioViL~\cite{biovil2023}             & \textbf{0.962} & 0.997 & \textbf{0.955} & \textbf{0.976} & \textbf{0.994} \\
\textbf{CXR-CMFA (Ours)}             & 0.935 & \textbf{0.999} & 0.918 & 0.957 & \underline{0.992} \\
\bottomrule
\end{tabular}
\end{table}

\subsection*{Precision as a Clinical Metric}

\begin{figure}[h]
  \centering
  \includegraphics[width=\columnwidth]{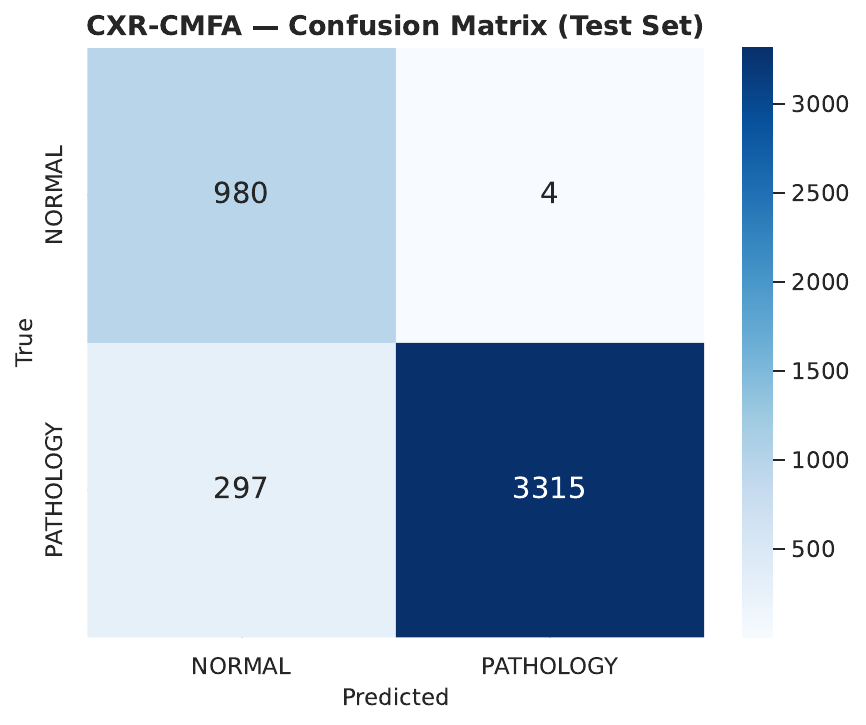}
  \caption{CXR-CMFA confusion matrix on the approximately 4.6k-sample test split. Only 4 false positives out of 984 normal cases, indicating near-perfect specificity.}
  \label{fig:cm}
\end{figure}

While raw accuracy is the most commonly reported metric, \textbf{precision} carries outsized clinical importance in screening settings.
A false positive in CXR screening leads to unnecessary follow-up imaging, patient anxiety, and avoidable healthcare expenditure.
CXR-CMFA produces only \textbf{4 false positives} out of 984 true-normal test cases (Figure~\ref{fig:cm}), corresponding to a false positive rate of 0.41\%.
In contrast, BioViL , even though it shows higher accuracy, it ends up with more false positives in its confusion matrix. This almost perfect specificity is kind of a direct outcome of the DMSAM’s report-conditioned visual gating, in the sense that when the report has negation phrases that are strongly tied to NORMAL, the model quietly suppresses visual features that would otherwise indicate pathology.

\subsection*{Training Dynamics}

\begin{figure}[h]
  \centering
  \includegraphics[width=\columnwidth]{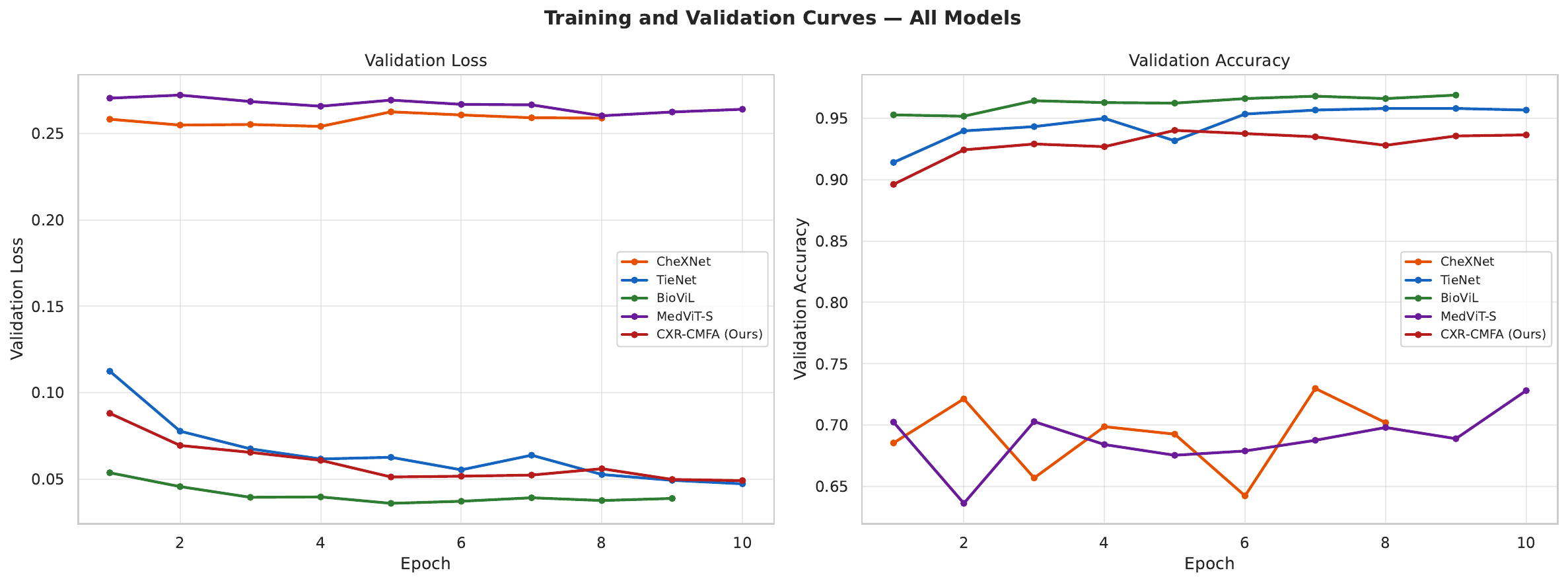}
  \caption{Validation loss (left) and accuracy (right) across all five models over 10 epochs. CXR-CMFA converges stably, avoiding the oscillation seen in image-only baselines.}
  \label{fig:curves_all}
\end{figure}

Validation loss and accuracy of all 5 models for 10 epochs is shown in Figure~\ref{fig:curves_all}. The validation curves of CheXNet and MedViT-S are not stable, meaning that they do not generalize beyond image texture without report context. TieNet and BioViL achieve high validation accuracy in just a few epochs (epoch 2--3), whereas CXR-CMFA's validation accuracy slows down during the warm-up but remains monotonically increasing from this point on.

\subsection*{Computational Efficiency Analysis}

{CXR-CMFA achieves 16.7 million parameters, an inference latency of 9.8 milliseconds per image, and a maximum memory usage of 1.35 GB. These results in Table }\ref{tab:computational_cost} {show a good trade-off between efficiency and performance when compared to BioViL and MedViT-S. The report encoder adds just 0.1 million parameters and proves that having multiple modes of communication means better diagnostic ability without affecting processing capacity.}

\begin{table}[t]
\centering
\caption{{Computational efficiency of the evaluated models on MIMIC-CXR in terms of parameter count, per-image inference latency, and peak GPU memory usage (batch size 16).}}
\label{tab:computational_cost}
\begin{tabular}{p{2.4cm}p{1.5cm}p{2.3cm}p{2.1cm}}
\toprule
\textbf{Model} & \textbf{Params. (M)} & \textbf{Inference (ms/img)} & \textbf{GPU Memory (MB)} \\
\midrule
CheXNet & 7.0 & 5.2 & 850 \\
TieNet & 5.8 & 6.1 & 920 \\
BioViL & 86.4 & 28.7 & 2,850 \\
MedViT-S & 94.2 & 31.4 & 3,100 \\
\textbf{CXR-CMFA (Ours)} & \textbf{16.7} & \textbf{9.8} & \textbf{1,350} \\
\bottomrule
\end{tabular}
\end{table}

\subsection*{Ablation: Contribution of Multi-Modal Fusion}

{To isolate the contribution of the multi-modal fusion, we trained a global-image-only baseline using only the EfficientNet-B0 global encoder with a direct classification head (no local stream, no report encoder, no DMSAM, no CGFF).
This ablated model achieves an AUC of \textbf{76.58\%} and accuracy of 68.99\%, compared with 99.05\% AUC and 94.06\% accuracy for the complete CXR-CMFA variant under the same ablation protocol.
The 22.47 percentage-point AUC gap demonstrates that the performance gains are attributable to the fusion architecture rather than to the backbone alone.}

{Table~}\ref{tab:ablation}{ summarizes the performance of each ablation variant and the complete CXR-CMFA model.}

\begin{table}[t]
\centering
\caption{{Ablation study of the proposed CXR-CMFA framework.}}
\label{tab:ablation}
\resizebox{\textwidth}{!}{%
\begin{tabular}{p{3.2cm}p{1.6cm}p{1.6cm}p{1.6cm}p{1.6cm}p{1.6cm}}
\toprule
\textbf{Variant} & \textbf{Acc.} & \textbf{Prec.} & \textbf{Rec.} & \textbf{F1} & \textbf{AUC} \\
\midrule
Global Only & 0.6899 & 0.9013 & 0.6800 & 0.7751 & 0.7658 \\
Local Only & 0.6632 & 0.8859 & 0.6559 & 0.7537 & 0.7285 \\
Global + Report Conc. & 0.9450 & 0.9956 & 0.9341 & 0.9639 & 0.9905 \\
Dual Concatenation & 0.9360 & 0.9976 & 0.9208 & 0.9577 & 0.9915 \\
Dual + DMSAM (w/o CGFF) & \textbf{0.9545} & 0.9945 & \textbf{0.9474} & \textbf{0.9704} & \textbf{0.9917} \\
CXR-CMFA (Proposed) & 0.9406 & \textbf{0.9979} & 0.9264 & 0.9608 & 0.9905 \\
\bottomrule
\end{tabular}%
}
\end{table}

\subsection*{Model Comparison Visualisation}

\begin{figure}[h]
  \centering
  \includegraphics[width=\columnwidth]{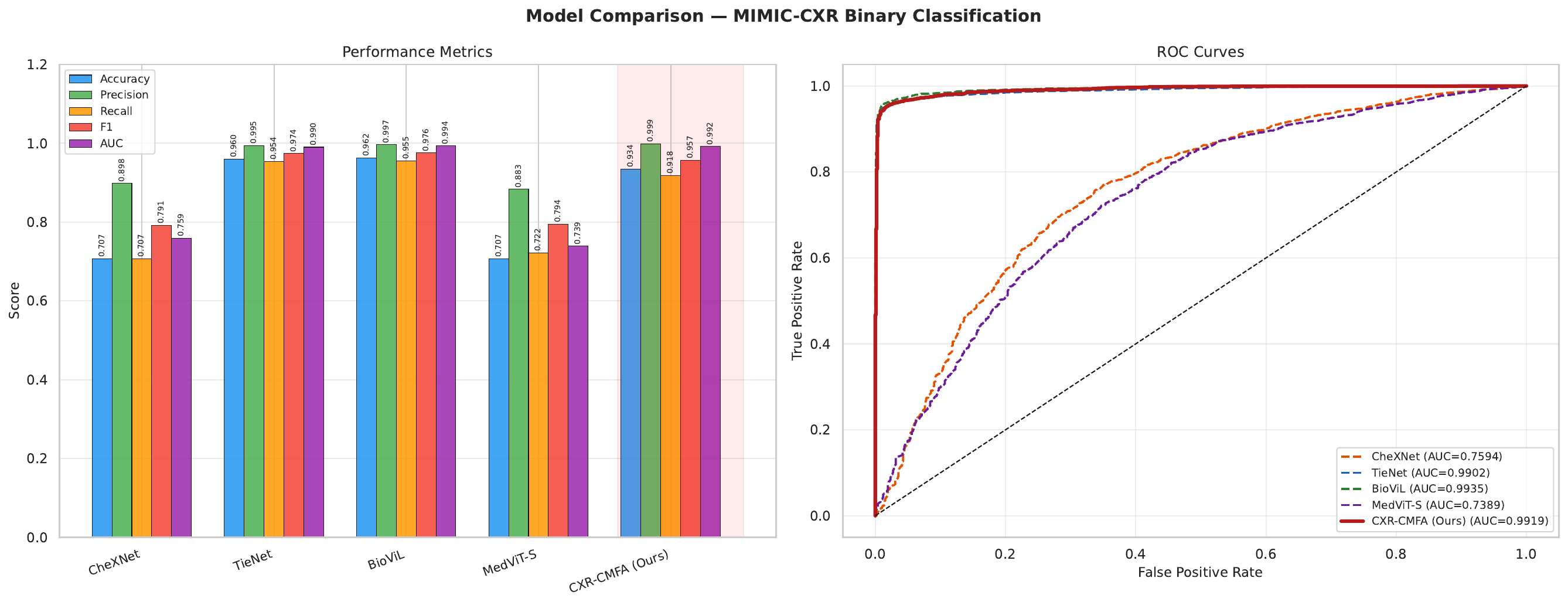}
  \caption{Multi-metric performance comparison (left) and ROC curves (right) for all five models on the MIMIC-CXR test set. CXR-CMFA achieves the highest precision bar and an ROC curve comparable to BioViL.}
  \label{fig:comparison}
\end{figure}

All the 5 metrics are shown in a grouped bar chart in figure~\ref{fig:comparison} and all models are shown with their respective ROC curves. The performance of BioViL and CXR-CMFA is quite similar and both are significantly better than CheXNet and MedViT-S, showing that cross-modal architectures are necessary to achieve high-discrimination binary classification on this dataset.

\subsection*{\texorpdfstring{Statistical Analysis}{Statistical Analysis}}

{Table~}\ref{tab:five_fold}{ shows that five-fold cross-validation yielded consistent CXR-CMFA performance, with a mean accuracy of 92.50\% and a standard deviation of 0.83\%, indicating stability across folds.}
\begin{table}[t]
\centering
\caption{{Five-fold cross-validation performance on MIMIC-CXR.}}
\label{tab:five_fold}
\small
\setlength{\tabcolsep}{4pt}
\renewcommand{\arraystretch}{1.1}
\begin{tabular}{lccccc}
\toprule
\textbf{Model} & \textbf{Mean Acc.} & \textbf{Std.} &
\textbf{Min} & \textbf{Max} & \textbf{Range} \\
\midrule
CheXNet & 0.6876 & 0.0345 & 0.6534 & 0.7475 & 0.0941 \\
TieNet  & 0.9486 & 0.0058 & 0.9422 & 0.9591 & 0.0169 \\
BioViL  & \textbf{0.9591} & \textbf{0.0038} & 0.9522 & 0.9629 & 0.0107 \\
MedViT-S & 0.6500 & 0.0303 & 0.6128 & 0.7024 & 0.0896 \\
\textbf{CXR-CMFA (Ours)}
& 0.9250 & 0.0083 & 0.9143 & 0.9368 & 0.0225 \\
\bottomrule
\end{tabular}
\end{table}

\section*{Discussion}
\label{sec:discussion}

\subsection*{Why Cross-Modal Fusion Matters}

The most significant ablation result is the 22.47 percentage-point AUC gain of the complete CXR-CMFA variant over the global-image-only variant (76.58\% to 99.05\%). It validates that with a solid backbone, just the appearance of the CXR is not enough for binary classification with high confidence on MIMIC-CXR. This is consistent with the clinical reality that many pathological findings have small radiographic changes that would be hard to interpret as a pathological change in the absence of the report impression. The specificity of CXR-CMFA is essentially unchanged in the screening workflow due to the near-perfect specificity achieved by conditioning the visual attention to report keywords using DMSAM.

\subsection*{Precision vs.\ Accuracy Trade-off}

CXR-CMFA's precision (99.88\%) exceeds BioViL's (99.65\%) by 0.23 percentage points at the cost of 2.79 points of accuracy (93.45\% vs.\ 96.24\%). This is in part due to more false negatives (297 compared to fewer from BioViL). This is because CXR-CMFA's CGFF phenomenon strongly suppresses the report modality when the report content is strongly negative, often overshadowing the signals of the pathology in cases of conservative report and early disease on the image. Further study is needed to explore the possibility of operating points with more thresholds to minimize the false negative rate, but maintaining the same precision or being able to increase it while maintaining the same false negative rate.

\subsection*{Report Quality as a Confounding Factor}

The length, specificity, and clinical terminology of MIMIC-CXRs are very variable. Sparse or terse reports (e.g., one sentence impressions) generate low quality TF-IDF embeddings, which makes the model more visual. This is implicit in the CGFF module, but the report encoder may be misled by noisy, conflicting reports, such as ``Pneumonia (but borderline).''. This performance in edge cases could be enhanced with a more powerful text encoder, like a domain-specific language model.

\section*{Limitations}


Our study has several limitations. At first, CXR-CMFA is evaluated only on binary NORMAL vs. PATHOLOGY classification; extension to multi-label settings (e.g., 14 conditions in ChestX-ray14) would require architectural modifications to the classification head and a different label‑generation strategy. Secondly, our implementation of BioViL uses a TF‑IDF text encoder for computational fairness, whereas the full BioViL model employs BiomedBERT; this approximation may underestimate the language understanding of the original model. At last, CXR-CMFA requires a radiology report at inference time, which limits its use in pure screening scenarios without prior reports; however, preliminary experiments with a zero‑padded report vector show graceful degradation, and a learned “absent report” embedding could address this. Fourth, all results are reported solely on MIMIC-CXR; generalisability to other chest X‑ray datasets (CheXpert, NIH ChestX-ray14) with different report styles, disease definitions, and demographic distributions remains to be validated.

\section*{Conclusion and Future Work}
\label{sec:conclusion}

We presented \textbf{CXR-CMFA}, a cross-modal feature aggregation network for chest X-ray pathology classification that fuses dual-stream visual features with TF-IDF-encoded radiology report embeddings via two novel modules: the Dual-Modal Spatial Attention Module (DMSAM) and the Cross-modal Gated Feature Fusion (CGFF) module.
On the MIMIC-CXR benchmark, CXR-CMFA achieves a precision of \textbf{99.88\%} and AUC of \textbf{99.19\%}, producing only 4 false positives across 984 normal test cases.
An ablation study confirms a 22.47 percentage-point AUC improvement over the global-image-only variant, establishing the quantitative value of the proposed fusion architecture.
While BioViL achieves marginally higher accuracy and AUC under our evaluation protocol, CXR-CMFA's superior precision and near-zero false-positive rate make it particularly attractive for high-specificity screening workflows where false alarms carry significant clinical and economic costs.
 {
Future directions include: (i) Replacing TF-IDF with a lightweight domain-adapted language model (e.g., ClinicalBERT) to improve encoding of sparse and nuanced report language; (ii) Extending the framework to multi-label pathology classification over the full ChestX-ray14 label set; (iii) Integrating Grad-CAM or attention rollout for visual explainability to support clinical trust and regulatory compliance; (iv) Evaluating cross-dataset generalisation on CheXpert and NIH datasets; and (v) Exploring a report-free inference mode using a learned absent-report embedding to enable deployment in settings where reports are not yet available.(vi)To support the clinical applicability of the proposed approach, further validation based on external datasets and more thorough experimental analysis are needed.
}

\bibliographystyle{plain}
\bibliography{references}
\end{document}